\documentclass{fairmeta}

\usepackage{amsmath,amssymb,amsfonts,amsthm}
\usepackage{bbm}
\usepackage{tabularx}
\usepackage{array}
\usepackage{enumitem}
\usepackage{algorithm}
\usepackage{algorithmic}
\usepackage{pifont}
\usepackage{minted}
\usepackage{tikz}
\usepackage{pgfplots}
\usepackage{silence}
\usepackage[toc,page]{appendix}

\tcbuselibrary{skins, breakable, listings, minted}
\pgfplotsset{compat=1.18}
\usetikzlibrary{
  arrows.meta, positioning, shapes.geometric, shapes.symbols,
  fit, calc, backgrounds, decorations.pathreplacing,
  shadows.blur, patterns
}
\setlist[itemize]{topsep=3pt, itemsep=2pt, parsep=0pt, partopsep=0pt}
\setlist[enumerate]{topsep=3pt, itemsep=2pt, parsep=0pt, partopsep=0pt}

\setminted{
  fontsize=\footnotesize,
  breaklines,
  breakanywhere,
  autogobble,
  linenos,
  numbersep=8pt,
  xleftmargin=10pt,
  frame=none
}

\definecolor{SynBlue}{HTML}{1F4E79}
\definecolor{SynBlueLight}{HTML}{EEF5FB}
\definecolor{SynGray}{HTML}{5B6573}
\definecolor{SynGrayLight}{HTML}{F6F8FA}
\definecolor{SynGreen}{HTML}{2E7D32}
\definecolor{SynOrange}{HTML}{B26A00}
\definecolor{SynBorder}{HTML}{D6DEE8}
\definecolor{hardred}{HTML}{C0392B}
\definecolor{softblue}{HTML}{2980B9}
\definecolor{goldwin}{HTML}{27AE60}
\definecolor{lossred}{HTML}{E74C3C}
\definecolor{neutralgray}{HTML}{7F8C8D}
\definecolor{accent}{HTML}{8E44AD}
\definecolor{cmdpocolor}{HTML}{E67E22}
\definecolor{dpocolor}{HTML}{3498DB}
\definecolor{manifoldbg}{HTML}{EBF5FB}
\definecolor{pipelinebg}{HTML}{FEF9E7}
\definecolor{verifybg}{HTML}{E8F8F5}
\definecolor{strikered}{HTML}{FADBD8}
\definecolor{insertgreen}{HTML}{D5F5E3}
\definecolor{rewritebg}{HTML}{EBF5FB}

\theoremstyle{remark}
\newtheorem{remark}{Remark}[section]
\newtheorem{definition}{Definition}
\newtheorem{hypothesis}{Hypothesis}

\newtcolorbox{appendixnote}[1][]{
  enhanced,
  breakable,
  colback=SynBlueLight,
  colframe=SynBlue,
  boxrule=0.8pt,
  arc=2mm,
  left=1.2mm,right=1.2mm,top=1mm,bottom=1mm,
  fonttitle=\bfseries,
  title=#1
}

\newtcblisting{promptbox}[2][]{
  enhanced,
  breakable,
  listing engine=minted,
  minted language=text,
  colback=SynBlueLight,
  colframe=SynBlue,
  boxrule=0.9pt,
  arc=2mm,
  left=1mm,right=1mm,top=1mm,bottom=1mm,
  title={#2},
  fonttitle=\bfseries,
  coltitle=black,
  attach boxed title to top left={xshift=2mm,yshift=-2mm},
  boxed title style={
    colback=SynBlue,
    colframe=SynBlue,
    sharp corners,
    boxrule=0pt,
    left=1mm,right=1mm,top=0.6mm,bottom=0.6mm
  },
  listing only,
  minted options={
    fontsize=\footnotesize,
    breaklines,
    breakanywhere,
    autogobble,
    linenos,
    numbersep=8pt
  },
  #1
}

\newtcblisting{jsonbox}[2][]{
  enhanced,
  breakable,
  listing engine=minted,
  minted language=json,
  colback=SynGrayLight,
  colframe=SynGray,
  boxrule=0.8pt,
  arc=2mm,
  left=1mm,right=1mm,top=1mm,bottom=1mm,
  title={#2},
  fonttitle=\bfseries,
  listing only,
  minted options={
    fontsize=\footnotesize,
    breaklines,
    autogobble,
    linenos
  },
  #1
}

\newtcblisting{codebox}[2][]{
  enhanced,
  breakable,
  listing engine=minted,
  minted language=python,
  colback=black!2,
  colframe=SynBorder,
  boxrule=0.8pt,
  arc=2mm,
  left=1mm,right=1mm,top=1mm,bottom=1mm,
  title={#2},
  fonttitle=\bfseries,
  listing only,
  minted options={
    fontsize=\footnotesize,
    breaklines,
    autogobble,
    linenos,
    style=friendly
  },
  #1
}

\title{CM-DPO: Constraint-Margin Direct Preference Optimization for LLM Planning}

\author[1]{Rabimba Karanjai}
\author[1]{Qun Gu}
\author[1]{Hemanth Hegadehalli Madhavarao}
\author[1]{Wenhuan Sun}
\author[1]{Xiaojiao Yu}
\author[1]{Suryabhan Singh Hada}
\author[1]{Libin N. George}
\author[1]{Uma Kona}
\author[1]{Richard Williamson}
\author[1]{Linsey Pang}
\author[1]{Prakhar Mehrotra}
\affiliation[1]{PayPal AI Lab}
\contribution{rkaranjai@paypal.com}
\date{October 2026}

\begin{document}

\abstract{Direct Preference Optimization (DPO) treats all constraint violations equally: a \$1 budget overshoot and a \$1,000 overshoot induce the same training signal. It is also susceptible to length and style bias when preference pairs come from different model families. We introduce Constraint-Margin DPO (CM-DPO), which replaces DPO's binary preference signal with a continuous margin derived from a deterministic symbolic verifier and scaled by violation severity. Hard and soft constraints are separated through a lexicographic objective, ensuring hard constraints are never traded off against preferences. To supply CM-DPO with bias-reduced training pairs, we generate preference data through procedurally generated constraint profiles (DCCG) and minimal-edit distillation from a reasoning teacher (RT-MED), within a framework we call SynPlan-R. On TravelPlanner, NaturalPlan, and out-of-distribution PlanBench, an 8B model fine-tuned with CM-DPO achieves 89.2\% pass rate and 93.4\% solve rate, matching multi-agent systems at $13\times$ lower latency while outperforming GPT-4o on unseen Blocksworld by 9.2 points.}

\maketitle

\section{Introduction}
\label{sec:intro}

The deployment of Large Language Models (LLMs) is increasingly shifting from open-ended conversational generation toward agentic workflows that require precise, long-horizon planning under complex constraints~\cite{valmeekam2023planbench, xie2024travelplanner}.
Real-world planning queries are rarely governed by a single homogeneous rule set; instead, they combine \emph{hard} functional constraints, such as budget ceilings, temporal non-overlap, and capacity limits, that must be satisfied without exception, with \emph{soft} user preferences, such as preferred carriers, dining cuisines, and proximity to landmarks, that should be optimized but may be relaxed.
A user requesting ``a weekend trip under \$500 that avoids red-eye flights, \emph{preferably} with boutique hotels'' expects the budget ceiling and flight timing to be inviolable, while treating hotel style as a best-effort preference. Standard LLMs often conflate these two categories, producing plans that are semantically fluent but either violate hard constraints or unnecessarily sacrifice soft-constraint preferences~\cite{xie2024travelplanner}.

We characterize this failure landscape as a \textbf{Planning Trilemma} among three constraints:
\begin{itemize}[leftmargin=*]
  \item \textbf{Accuracy:} strict, verifiable satisfaction of all hard constraints and best-effort adherence to soft preferences.
  \item \textbf{Efficiency:} sub-second, single-pass inference compatible with real-time deployment.
  \item \textbf{Controllability:} responsiveness to dynamic, per-query constraint profiles without retraining.
\end{itemize}

Existing methods typically satisfy at most two of these three criteria. We propose a unified data-synthesis and alignment framework with the following contributions:

\begin{enumerate}[leftmargin=*]
  \item \textbf{Dynamic Contextual Constraint Graphs (DCCG):} A formal representation that decomposes planning problems into parameterized \emph{hard} and \emph{soft} constraint profiles over procedurally generated knowledge bases (synthetic flight and hotel databases with randomized prices and schedules). DCCG reduces the risk of domain memorization and encourages the model to learn transferable combinatorial reasoning (\S\ref{sec:dccg}).

  \item \textbf{Reasoning-Trace Minimal-Edit Distillation (RT-MED):} A teacher--student synthesis pipeline in which a reasoning model (DeepSeek-R1, o3-mini) corrects a flawed student draft through the \emph{minimum necessary semantic edits}, isolating the logical ``gradient'' from stylistic noise and mitigating the length bias inherent in full-rewrite distillation (\S\ref{sec:rtmed}).

  \item \textbf{Constraint-Margin Preference Optimization (CM-DPO):} An extension of DPO that injects a \emph{continuous} penalty margin into the preference loss, scaled by the deterministic severity of hard-constraint violations computed by a symbolic verifier. Soft-constraint adherence is incorporated as a lexicographic secondary objective, ensuring that hard constraints are never traded off against preference satisfaction (\S\ref{sec:cmdpo}).
\end{enumerate}

\section{Related Work}

\subsection{LLM Planning and Constraint Satisfaction}
\label{sec:rw-planning}

Open benchmarks have consistently documented the difficulty that LLMs face on constraint-heavy planning tasks. TravelPlanner~\cite{xie2024travelplanner} reports extremely low hard-constraint pass rates for open-source models on multi-day itinerary construction, while NaturalPlan~\cite{zheng2024naturalplan} shows similar weaknesses in temporal scheduling. PlanBench~\cite{valmeekam2023planbench} further shows that even frontier models struggle with classical Blocksworld-style domains, suggesting that planning failures are not solely a matter of scale, but also reflect limitations in representation and reasoning.

To address this gap, inference-time scaling methods interleave generation with environmental feedback. ReAct~\cite{yao2023react} augments generation with tool use, and Reflexion~\cite{shinn2023reflexion} adds self-critique, while LRPlan~\cite{karthikeyan2025lrplan} extends this approach to a full Multi-Agent System (MAS) in which a Large Reasoning Model (LRM) iteratively refines the output of a base LLM. Although effective, these methods trade \emph{efficiency} for \emph{accuracy}: latencies of 20--30\,s per query and multiplicative API costs limit their suitability for real-time deployment. A parallel line of work combines LLMs with external symbolic solvers~\cite{liu2023llmp, kambhampati2024llmmodulo}, which can provide formal correctness guarantees but often require domain-specific planner specifications, reducing generality. Constrained decoding methods~\cite{willard2023outlines, beurer2023prompting} enforce token-level grammatical constraints at generation time, but they are less suited to \emph{semantic} constraints whose validity depends on the global content of the plan (e.g., total cost $\leq$ budget).

\subsection{Preference Optimization and Its Limitations}
\label{sec:rw-dpo}

Direct Preference Optimization (DPO)~\cite{rafailov2024direct} simplified RLHF by removing the need for an explicit reward model and optimizing a closed-form objective over preference pairs. Subsequent work has identified two limitations that are especially relevant to constraint-aware planning.

Standard DPO assigns the same implicit reward gap to all rejected samples, regardless of the severity of the failure. Identity Preference Optimization (IPO)~\cite{azar2024general} addresses overfitting through a regularized objective, but it does not encode domain-specific violation semantics. Kahneman--Tversky Optimization (KTO)~\cite{ethayarajh2024kto} replaces pairwise preferences with per-example utility, but it likewise does not distinguish among different kinds of negative outcomes. SMAUG-DPO~\cite{pal2024smaug} introduces a pairwise margin, but the margin is derived from reward-model scores rather than deterministic constraint verification.

When chosen and rejected responses come from different model families, DPO can rely on superficial distributional features rather than the underlying correction itself~\cite{park2024disentangling, singhal2024long}.
This concern is closely related to the broader observation that DPO can be sensitive to response length and may over-reward stylistic cues over substantive quality. Configurable Preference Tuning (CPT)~\cite{gallego2025configurable} mitigates this issue only partially, since its rubrics are primarily stylistic rather than logical.

CM-DPO differs from these approaches in two respects: it derives its margin from a \emph{deterministic symbolic verifier} rather than a learned reward signal, and it uses a \emph{lexicographic} objective that prioritizes hard-constraint satisfaction over soft-preference adherence (\S\ref{sec:cmdpo}).

\subsection{Synthetic Data and Preference Alignment}

Synthetic data has become an increasingly important tool for model alignment. Self-Instruct~\cite{wang2023selfinstruct} showed that models can generate useful instruction-tuning data for themselves, and DPO~\cite{rafailov2024direct} removed the need for explicit reward models by optimizing directly over preference pairs. Configurable Preference Tuning (CPT)~\cite{gallego2025configurable} extended this setup by conditioning DPO on varying ``rubrics.'' However, standard DPO can become brittle when negative samples are distinguishable through superficial stylistic traits rather than substantive errors. SynPlan-R addresses this issue by generating counterfactual minimal edits and by modifying the DPO objective to account for continuous constraint-violation margins.

\subsection{Knowledge Distillation for Reasoning}
\label{sec:rw-distill}

Distilling reasoning ability from larger teacher models into smaller ones has received renewed attention following the success of chain-of-thought prompting~\cite{wei2022chain}. Distilling Step-by-Step~\cite{hsieh2023distilling} uses rationales as intermediate supervision, and more recent work has explored distilling extended reasoning traces from open reasoning models such as DeepSeek-R1~\cite{deepseekai2025r1, yu2024distilling}. Most of the literature focuses on broad reasoning domains such as mathematics and code generation. SynPlan-R instead targets \emph{constraint-satisfaction reasoning} and introduces RT-MED to reduce the tendency of the student to imitate the teacher's verbose reasoning style rather than internalize the underlying logical correction.

\section{Dynamic Contextual Constraint Graphs}
\label{sec:dccg}

We represent each planning instance as a \emph{constraint profile} that separates hard constraints, soft preferences, and the local knowledge base used for verification. This formulation makes the distinction between feasibility and preference explicit and provides a common interface across domains.

\begin{definition}[Constraint Profile]
  \label{def:profile}
  A profile $\mathbf{p} \in \Omega$ is a tuple
  $\langle \mathbf{C}_{hard}, \mathbf{C}_{soft},
  \mathcal{K}_{local} \rangle$, where:
  \begin{itemize}[nosep]
    \item $\mathbf{C}_{hard} = \{c_1, \dots, c_m\}$ is the set of hard
      constraints, evaluated by a Boolean verifier
      $\mathcal{V}_{hard}: \mathcal{Y} \times
      \mathbf{C}_{hard} \times \mathcal{K}_{local}
      \to \{0, 1\}^m$.
    \item $\mathbf{C}_{soft} = \{c_{m+1}, \dots, c_n\}$ is the set of
      soft preferences, scored by
      $\mathcal{S}_{soft}: \mathcal{Y} \times
      \mathbf{C}_{soft} \times \mathcal{K}_{local}
      \to [0, 1]^{n-m}$.
    \item $\mathcal{K}_{local}$ is a procedurally generated local knowledge base provided in context, requiring the model to reason over instance-specific facts rather than rely on memorized domain knowledge.
  \end{itemize}
\end{definition}

Candidate plans are ranked first by hard-constraint satisfaction and only then, among plans with equal hard-constraint status, by soft-preference score. Let $H(y) = \sum_j \mathcal{V}_{hard,j}(y)$. We write $y \succ y'$ iff $H(y) > H(y')$, or $H(y) = H(y')$ and $\bar{\mathcal{S}}(y) > \bar{\mathcal{S}}(y')$.

\subsection{Domain Generalization}

We compare the framework in two domains, travel planning and retail recommendation, each with its own constraint profile and verifier.

\paragraph{Travel:}
We can formalize a travel planning as
\[
\mathbf{p}_{travel} =
\langle
\underbrace{\beta, \tau, \eta}_{\mathbf{C}_{hard}},
\underbrace{\Lambda}_{\mathbf{C}_{soft}},
\mathcal{K}_{local}
\rangle,
\]
where \(\beta\) is the budget, \(\tau\) the temporal window, \(\eta\) the minimum pace requirement, and \(\Lambda\) the set of preferred attributes. Hard constraints are verified by
\begin{equation}
  \mathcal{V}_{cost}(y) = \mathbbm 1_{\sum_{i \in \mathcal{I}_y} \text{Price}(i, \mathcal{K}) \leq \beta}, \label{eq:v-cost}
\end{equation}
and
\begin{equation}
  \mathcal{V}_{time}(y) = \bigwedge_i \mathbbm 1_{t_{arr}(v_{i+1}) - t_{dep}(v_i) \geq d(v_i, v_{i+1})}. \label{eq:v-time}
\end{equation}
Soft-preference satisfaction is defined as
\[
\mathcal{S}_{attr}(y, \Lambda) =
\frac{\left|\{v \in \mathcal{I}_y :
\text{Attrs}(v) \cap \Lambda \neq \emptyset\}\right|}
{|\mathcal{I}_y|}.
\]
For cost violations, we define the continuous severity as
\begin{equation}
  \label{eq:delta-cost}
  \Delta\mathcal{V}_{cost}(y) = 
  \left(\sum_i \text{Price}(i, \mathcal{K}) - \beta \right)^+,
\end{equation}
where $(\cdot)^+=\max(0,\cdot)$.

\paragraph{Retail:}
We formalize a retail product search as
\[
\mathbf{p}_{retail} =
\langle
\underbrace{\kappa, \mathcal{F}_{req}, \mathcal{G}_{compat}}_{\mathbf{C}_{hard}},
\underbrace{\mathcal{R}_{pref}}_{\mathbf{C}_{soft}},
\mathcal{K}_{local}
\rangle,
\]
where \(\kappa\) is the budget, \(\mathcal{F}_{req}\) the set of required features, \(\mathcal{G}_{compat}\) a compatibility graph, and \(\mathcal{R}_{pref}\) the preferred brands or retailers.

A deterministic function \(\mathcal{T}: \Omega \to \Sigma^*\) maps each constraint profile to a natural-language system prompt \(s\), which is then used to condition generation. This projection allows the same underlying profile representation to be expressed as a natural-language planning prompt without changing the verifier-defined semantics.

\section{The SynPlan-R Synthesis Pipeline}
\label{sec:pipeline}

\begin{figure}
    \centering
    \includegraphics[width=0.84\linewidth]{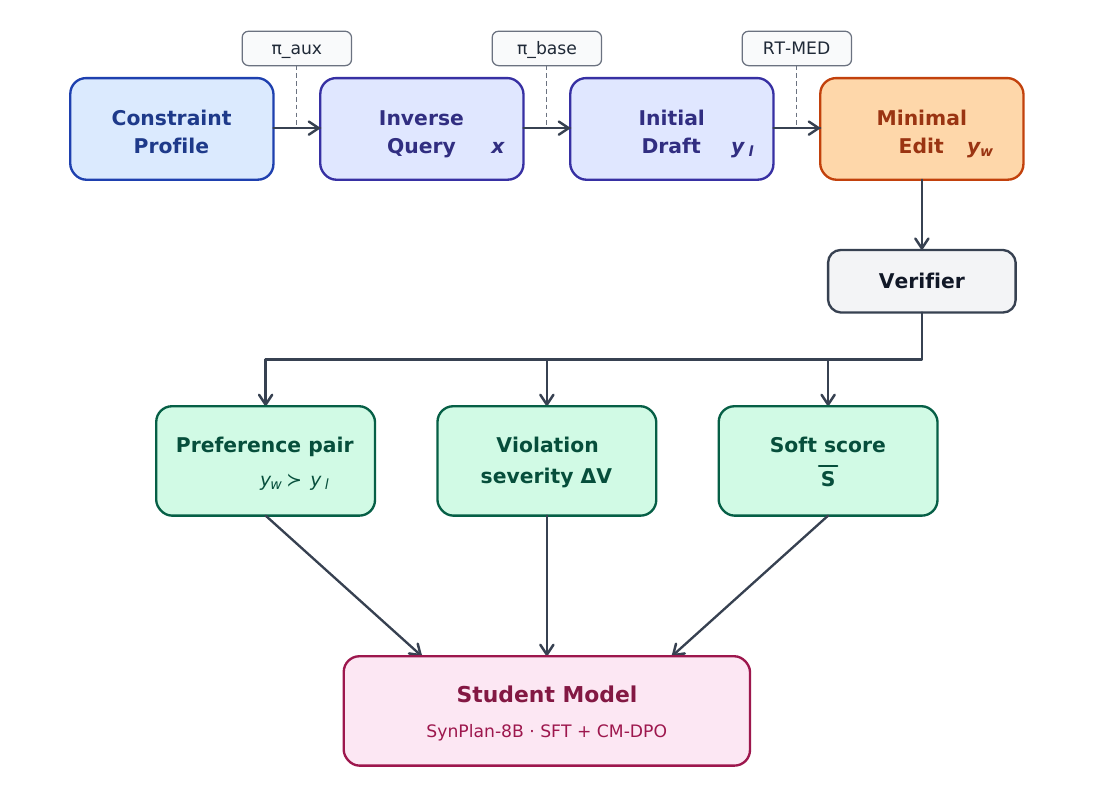}
    \caption{\textbf{The SynPlan-R architecture.} Phase~1: DCCG decomposes each profile into hard constraints $\mathbf{C}_{hard}$ (red), soft preferences $\mathbf{C}_{soft}$ (blue), and a procedural knowledge}
    \label{fig:architecture-phases}
\end{figure}

The pipeline constructs a set of verified preference pairs,
\[
\mathcal{D} = \{(s_i, x_i, y_{w,i}, y_{l,i},
\Delta\mathcal{V}_i, \bar{\mathcal{S}}_i)\}_{i=1}^N,
\]
in four phases (Figure~\ref{fig:architecture-phases}).

\paragraph{Phase 1: Inverse Constraint Querying.}
\label{sec:inv-query}
An auxiliary policy $\pi_{aux}$ generates implicit user queries,
\[
x \sim \pi_{aux}(x \mid \text{Prompt}_{inv} \oplus \mathbf{p}),
\]
yielding requests such as ``affordable weekend getaway'' for a $\$600$ budget profile. This setup requires the downstream model to infer the relevant constraints from \(s\) rather than relying on an explicitly stated rule list.

\paragraph{Phase 2: Baseline Draft ($y_l$).}
\label{sec:baseline}
A student policy $\pi_{base}$ (GPT-4o-mini) generates an initial draft,
\[
y_l \sim \pi_{base}(y \mid s, x).
\]

\begin{hypothesis}[Heuristic Failure]
  \label{hyp:failure}
  Standard policies often approximate hard constraints using surface-level heuristics, which lead to near-miss violations ($\text{Cost}(y_l) \approx \beta$ but $\text{Cost}(y_l) > \beta$). These near-misses provide especially informative negative samples.
\end{hypothesis}

\paragraph{Phase 3: RT-MED.}
\label{sec:rtmed}
To reduce the extent to which DPO can exploit stylistic differences between teacher and student outputs, we use \textbf{Reasoning-Trace Minimal-Edit Distillation (RT-MED)}. The teacher model $\pi_{ref}$ (o3-mini) receives $y_l$, $s$, and $\mathcal{K}_{local}$, and produces
\begin{equation}
  \label{eq:rtmed}
  y_w = \texttt{<think>} \oplus c_{trace} \oplus
  \texttt{</think>} \oplus \text{Edit}(y_l),
\end{equation}
where $Edit(y_l)$ is heuristically guided by a system prompt instructing the teacher to isolate and resolve the violation. Rather than performing a formal combinatorial search or strict mathematical optimization over semantic distance, the teacher is prompted to greedily swap only the specific entities responsible for the violation while leaving the surrounding valid sub-plans and formatting untouched. In practice, the teacher typically changes only the entities responsible for the violation (\(\sim 32\) tokens versus \(\sim 245\) for full rewrites), which helps separate constraint-relevant corrections from stylistic variation (Table~\ref{tab:ablation}, Figure~\ref{fig:rtmed-example}).

\begin{figure*}[t]
  \centering
  \begin{minipage}{.55\textwidth}
  \resizebox{\textwidth}{!}{%
\begin{tikzpicture}[
    >=Stealth,
    every node/.style={font=\small}
]
\begin{scope}[on background layer]
    \fill[lossred!5, rounded corners=5pt] (0, 0.7) rectangle (4.8, 8);
\end{scope}
\node[font=\small\bfseries, text=lossred] at (2.4, 7.7)
    {Student Draft $y_l$ ($\pi_{base}$)};
\draw[lossred!40, line width=0.5pt] (0.3, 7.45) -- (4.5, 7.45);
\node[anchor=north west, font=\scriptsize, text=black!80, align=left]
at (0.4, 7.3)
{
    \textbf{Day 1:} Fly SFO $\to$ NYC\\[2pt]
    \qquad United \#482, dep 8:00am\\[2pt]
    \qquad $\mathcal{K}$: \$180
};

\fill[anchor=north west, strikered, rounded corners=2pt]
    (0.2, 5.55) rectangle (4.6, 6.23);
\node[anchor=north west, font=\scriptsize\bfseries, align=left,
text=hardred] at (0.4, 6.3)
{
    \textbf{Hotel}: The Standard NYC\\[2pt]
    \qquad \$195/night $\times$ 2 = \$390
};

\node[anchor=north west, font=\scriptsize, text=black!80, align=left]
at (0.4, 5.6)
{
    \textbf{Day 2:} Central Park, MoMA\\[2pt]
    \qquad Park: free \quad MoMA: \$30\\[2pt]
    \textbf{Dinner:} Le Bernardin\\[2pt]
    \qquad \$85 per person\\[2pt]
    \textbf{Day 3:} Brooklyn Bridge, fly back\\[2pt]
    \qquad Return flight: \$165
};

\draw[lossred!40, line width=0.5pt] (0.3, 3.35) -- (4.5, 3.35);
\fill[strikered, rounded corners=2pt]
    (0.2, 2.85) rectangle (4.6, 3.25);
\node[anchor=north west, font=\scriptsize\bfseries, text=hardred] at (0.4, 3.3)
    {Total: \$650 \qquad \ding{55} OVER BUDGET};
\node[anchor=north west, font=\scriptsize, text=hardred] at (0.4, 2.8)
    {$\Delta\mathcal{V}_{cost} = $ \$650 $-$ \$600 $=$ \$50};

\fill[softblue!10, rounded corners=2pt] (0.2, 1.8) rectangle (4.6, 2.3);
\node[anchor=north west, font=\scriptsize, text=softblue] at (0.4, 2.3)
    {Soft: ``The Standard'' = boutique \checkmark};
\node[font=\scriptsize\bfseries, text=lossred, fill=lossred!10,
      rounded corners=3pt, inner sep=4pt] at (2.4, 1.1)
    {$\mathcal{V}_{hard}(y_l) = 0$ \quad REJECTED};

\begin{scope}[on background layer]
    \fill[goldwin!5, rounded corners=5pt] (5, 0.7) rectangle (9.8, 8);
\end{scope}
\node[font=\small\bfseries, text=goldwin] at (7.4, 7.7)
    {RT-MED Edit $y_w$ ($\pi_{ref}$)};
\draw[goldwin!40, line width=0.5pt] (5.3, 7.45) -- (9.5, 7.45);
\node[anchor=north west, font=\scriptsize, align=left,
text=black!80] at (5.4, 7.3)
    {
        \textbf{Day 1:} Fly SFO $\to$ NYC\\[2pt]
        \qquad United \#482, dep 8:00am\\[2pt]
        \qquad $\mathcal{K}$: \$180
    };
\fill[anchor=north west, insertgreen, rounded corners=2pt]
    (5.2, 5.55) rectangle (9.6, 6.23);
\node[anchor=north west, font=\scriptsize\bfseries, align=left,
text=goldwin] at (5.4, 6.3)
{
    \textbf{Hotel}: Hampton Inn Midtown\\[2pt]
    \qquad \$120/night $\times$ 2 = \$240
};

\node[anchor=north west, font=\scriptsize, text=black!80, align=left]
at (5.4, 5.6)
{
    \textbf{Day 2:} Central Park, MoMA\\[2pt]
    \qquad Park: free \quad MoMA: \$30\\[2pt]
    \textbf{Dinner:} Le Bernardin\\[2pt]
    \qquad \$85 per person\\[2pt]
    \textbf{Day 3:} Brooklyn Bridge, fly back\\[2pt]
    \qquad Return flight: \$165
};

\draw[goldwin!40, line width=0.5pt] (5.3, 3.35) -- (9.5, 3.35);

\fill[insertgreen, rounded corners=2pt]
    (5.2, 2.85) rectangle (9.6, 3.25);
\node[anchor=north west, font=\scriptsize\bfseries, text=goldwin] at (5.4, 3.3)
    {Total: \$550 \qquad \qquad \ding{51} UNDER};
\node[anchor=north west, font=\scriptsize, text=goldwin] at (5.4, 2.8)
    {Slack: \$600 $-$ \$550 $=$ \$50};
\fill[softblue!10, rounded corners=2pt] (5.2, 1.8) rectangle (9.6, 2.3);
\node[anchor=north west, font=\scriptsize, text=neutralgray] at (5.4, 2.3)
    {Soft: Hampton Inn $\neq$ boutique};

\node[font=\scriptsize\bfseries, text=goldwin, fill=goldwin!10,
      rounded corners=3pt, inner sep=4pt] at (7.4, 1.1)
    {$\mathcal{V}_{hard}(y_w) = 1$ \quad ACCEPTED};

\begin{scope}[on background layer]
    \fill[rewritebg, rounded corners=5pt] (10, 0.7) rectangle (15, 8);
\end{scope}
\node[font=\small\bfseries, text=neutralgray!70!black] at (12.5, 7.7)
    {Full Rewrite $y_w'$};
\draw[neutralgray!30, line width=0.5pt] (10.3, 7.45) -- (14.7, 7.45);
\node[anchor=north west, font=\scriptsize, text=neutralgray!70!black, align=left] at (10.4, 7.3)
    {
        \textbf{Day 1:} Fly JFK via Spirit\\[2pt]
        \qquad {\it // different airline, route}\\[2pt]
        \qquad ~
    };
\node[anchor=north west, font=\scriptsize, align=left,
    text=neutralgray!70!black] at (10.4, 6.3)
{
    \textbf{Hotel}: Pod 51 NYC\\[2pt]
    \qquad \textit{// different hotel, structure} \\[2pt]
};

\node[anchor=north west, font=\scriptsize, align=left,
    text=neutralgray!70!black] at (10.4, 5.6)
{
    \textbf{Day 2:} High Line, Chelsea\\[2pt]
    \qquad \textit{// different activities} \\[2pt]
    \textbf{Dinner:} Joe's Pizza\\[2pt]
    \qquad \textit{// different restaurant}\\[2pt]
    \textbf{Day 3:} DUMBO, return\\[2pt]
    \qquad \textit{// different itinerary}
};
\draw[neutralgray!30, line width=0.5pt] (10.3, 3.35) -- (14.5, 3.35);

\node[anchor=north west, font=\scriptsize\bfseries, text=neutralgray!70!black] at (10.4, 3.3)
    {Total: \$420 \qquad \qquad \ding{51}};
\fill[lossred!10, rounded corners=2pt] (10.2, 1.5) rectangle (14.6, 2.8);
\node[anchor=north west, font=\scriptsize, text=lossred!80!black,
      text width=3.4cm, align=left] at (10.4, 2.8) {
    \textbf{Problem:} DPO sees $\sim$245 token delta.
    Student learns to mimic teacher's \textit{vocabulary}
    and \textit{style}, not the constraint logic.
};
\node[font=\scriptsize\bfseries, text=neutralgray, fill=neutralgray!10,
      rounded corners=3pt, inner sep=4pt] at (12.4, 1.1)
    {Style bias $\to$ 65\% PR};

\draw[black!20, line width=0.3pt] (0, 0.5) -- (15, 0.5);

\begin{axis}[
    at={(15mm, 3mm)},
    anchor=north west,
    xbar,
    width=7cm,
    height=3cm,
    xlabel={Tokens},
    ytick={(0,1)},
    yticklabels={$y_w'$, $y_w$},
    xmin=0,
    nodes near coords,
    enlarge x limits={upper, value=0.2},
    ymin=-1,
    ymax=2,
    legend style={at={(0.95,0.05)}, anchor=south east},
    /pgf/bar shift={0pt},
]
\addplot [fill=rewritebg, draw=neutralgray] coordinates {(245,0)};
\addplot [fill=goldwin!50, draw=goldwin] coordinates {(32,1)};

\end{axis}

\begin{axis}[
    at={(8cm, 3mm)},
    anchor=north west,
    xbar,
    width=7cm,
    height=3cm,
    xlabel={PR},
    ytick=data,
    yticklabels={},
    xmin=0,
    nodes near coords,
    enlarge x limits={upper, value=0.3},
    enlarge y limits=1,
    legend style={at={(0.95,0.05)}, anchor=south east},
    /pgf/bar shift={0pt},
    point meta={x*100},
    xticklabel={\pgfmathparse{\tick*100}\pgfmathprintnumber{\pgfmathresult}\%},
    nodes near coords={\pgfmathprintnumber\pgfplotspointmeta\%},
    nodes near coords align={horizontal},
]
\addplot [fill=rewritebg, draw=neutralgray] coordinates {(0.892,0)};
\addplot [fill=goldwin!50, draw=goldwin] coordinates {(0.652,1)};
\end{axis}

\end{tikzpicture}%
}% end resizebox
  \end{minipage}
  \hfill
  \begin{minipage}{.4\textwidth}
  \caption{\textbf{RT-MED qualitative example} The hard constraint is total cost $\leq 600$, and the soft constraint is a preference for boutique hotels. Left: the student draft violates the budget constraint (hotel at \$195/night, yielding a total cost of \$650). Center: RT-MED swaps only the hotel (\(\sim 12\) tokens). Right: a full rewrite changes \(\sim 245\) tokens and introduces style bias. Bottom: token-count and PR comparison.}
  \label{fig:rtmed-example}      
  \end{minipage}
\end{figure*}

\paragraph{Phase 4: Verification and Severity.}
\label{sec:verification}
A symbolic verifier $\mathcal{V}$ evaluates each plan against $\mathcal{K}_{local}$. A pair is included only if $\mathcal{V}_{hard}(y_w) = \mathbf{1}$ and there exists at least one constraint \(j\) such that $\mathcal{V}_{hard,j}(y_l) = 0$. The verifier then records the continuous violation severity $\Delta\mathcal{V}(y_l)$ and the soft-preference score $\bar{\mathcal{S}}_{soft}(y_w)$, as defined in Eq.~\ref{eq:delta-cost}.

\begin{minipage}[t]{.45\textwidth}
\begin{algorithm}[H]
  \caption{SynPlan-R Synthesis with RT-MED}
  \label{alg:synplan}
  \begin{algorithmic}[1]
    \REQUIRE $\Omega$, $\pi_{base}$, $\pi_{ref}$, $\pi_{aux}$
    \ENSURE $\mathcal{D}$
    \STATE $\mathcal{D} \leftarrow \emptyset$
    \FOR{$\mathbf{p} \sim \Omega$}
    \STATE $s, \mathcal{K} \leftarrow \mathcal{T}(\mathbf{p})$;
    $x \leftarrow \pi_{aux}(\cdot \mid \mathbf{p})$
    \STATE $y_l \leftarrow \pi_{base}(y \mid s, x)$
    \STATE $y_w \leftarrow \pi_{ref}(\text{RT-MED} \mid s, x, y_l, \mathcal{K})$
    \STATE Compute $\mathcal{V}_{hard}$, $\Delta\mathcal{V}$,
    $\bar{\mathcal{S}}$
    \IF{$\mathcal{V}_{hard}(y_w)\!=\!\mathbf{1} \land \exists
    j\!:\!\mathcal{V}_{hard,j}(y_l)\!=\!0$}
    \STATE $\mathcal{D} \leftarrow \mathcal{D} \cup \{(s, x, y_w,
    y_l, \Delta\mathcal{V}, \bar{\mathcal{S}})\}$
    \ENDIF
    \ENDFOR
  \end{algorithmic}
\end{algorithm}
\end{minipage}
\hfill
\begin{minipage}[t]{.5\textwidth}
\begin{algorithm}[H]
\caption{CM-DPO Alignment (Stage 2)}
\begin{algorithmic}[1]
\REQUIRE SFT-init, frozen, dataset $(\pi_{\theta}, \pi_{ref}, \mathcal{D})$
\WHILE{not converged}
    \STATE Sample $(s, x, y_w, y_l, \Delta\tilde{\mathcal{V}}, \overline{\mathcal{S}}) \sim \mathcal{D}$
    \STATE $m_c \leftarrow \tanh(\lambda \cdot \Delta\tilde{\mathcal{V}})$
    \STATE $\Delta_{\theta} \leftarrow \beta_d \left[ \log \frac{\pi_{\theta}(y_w|x)}{\pi_{ref}(y_w|x)} - \log \frac{\pi_{\theta}(y_l|x)}{\pi_{ref}(y_l|x)} \right]$
    \STATE $\mathcal{L}_{hard} \leftarrow -\log \sigma(\Delta_{\theta} - \gamma \cdot m_c)$
    \STATE $\mathcal{L}_{soft} \leftarrow -\log \sigma(\Delta_{\theta}) \cdot (\overline{\mathcal{S}}(y_w) - \overline{\mathcal{S}}(y_l))^{+}$
    \STATE $\mathcal{L}_{CM} \leftarrow \mathcal{L}_{hard} + \alpha \cdot \mathbbm{1}_{\mathcal{V}_{hard}(y_w) = 1} \cdot \mathcal{L}_{soft}$
    \STATE Update $\theta$ via $\nabla_{\theta} \mathcal{L}_{CM}$
\ENDWHILE
\end{algorithmic}
\end{algorithm}
\end{minipage}

\section{Constraint-Margin Preference Optimization}
\label{sec:cmdpo}

\begin{figure}[t]
  \centering
  \begin{minipage}{.55\textwidth}
  \includegraphics[width=\textwidth]{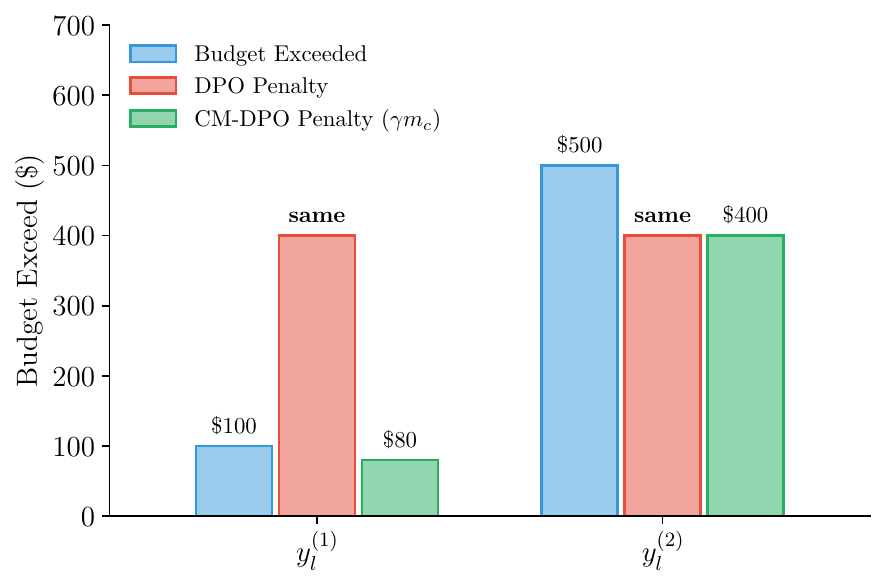}
  \end{minipage}
  \hfill
  \begin{minipage}{.4\textwidth}
  \caption{\textbf{CM-DPO vs.\ Standard DPO.} Standard DPO assigns the same penalty regardless of violation severity. CM-DPO scales the margin $\gamma m_c$ proportionally to the budget overshoot, producing stronger repulsion for larger violations.}
  \label{fig:cmdpo}
  \end{minipage}
\end{figure}

We begin with standard DPO and then extend it in two steps. First, we replace the binary preference signal with a continuous margin derived from the severity of hard-constraint violations. Second, we add a separate soft-constraint objective that is activated only when hard constraints are satisfied.

\subsection{Hard-Constraint Margin}
\label{sec:hard-margin}

Standard DPO treats all rejected responses equally, regardless of the severity of the violation. In constraint-aware planning, however, a small budget overshoot and a catastrophic failure should not induce the same update. To encode this distinction, we define a continuous margin
\[
m_c = \tanh(\lambda \cdot \Delta\mathcal{V}(y_l)) \in [0,1],
\]
where \(\Delta\mathcal{V}(y_l)\) is the verifier-computed severity of the rejected plan and \(\lambda > 0\) controls sensitivity. The \(\tanh\) transformation keeps the margin bounded while remaining approximately linear near the origin.

We then modify the hard-constraint objective as
\begin{equation}
  \label{eq:cmdpo-hard}
  \mathcal{L}_{\text{hard}} = -\mathbb{E}\left[
  \log \sigma\left(\Delta_\theta - \gamma \cdot m_c\right) \right],
\end{equation}
where \(\gamma\) controls how strongly violation severity shifts the decision boundary. Larger violations therefore induce a larger effective margin and a stronger corrective update.

\begin{remark}[Gradient amplification]
  \label{rem:gradient}
  Let \(z = \Delta_\theta - \gamma m_c\). The gradient term includes \(\sigma(-z)\), so increasing \(m_c\) pushes \(z\) downward and increases the update magnitude. Intuitively, the model is pushed more strongly away from catastrophic violations while still receiving smoother corrective feedback near the feasibility boundary.
\end{remark}

\subsection{Soft-Constraint Objective}
\label{sec:soft-loss}

Hard feasibility alone is not sufficient. Among feasible plans, the model should also prefer those that better satisfy soft constraints. We therefore introduce a secondary objective
\begin{equation}
  \label{eq:soft-loss}
  \mathcal{L}_{\text{soft}} = -\log \sigma(\Delta_\theta)
  \cdot \left(\bar{\mathcal{S}}(y_w) -
  \bar{\mathcal{S}}(y_l)\right)^{+}.
\end{equation}
This term is active only when the chosen response improves soft-preference satisfaction. If the chosen plan gives up a soft preference in order to restore hard feasibility, the soft term contributes zero rather than penalizing that correction.

\subsection{Combined Objective}

We combine the two terms as
\begin{equation}
  \label{eq:cmdpo-full}
  \mathcal{L}_{\text{CM}} =
  \mathcal{L}_{\text{hard}}(m_c)
  + \alpha \cdot
  \mathbbm{1}[\mathcal{V}_{hard}(y_w) = \mathbf{1}]
  \cdot \mathcal{L}_{\text{soft}},
\end{equation}
where \(\alpha \geq 0\) controls the weight of the soft-constraint objective. The indicator gate ensures that soft-preference optimization is applied only when the chosen plan is hard-feasible. In this sense, the objective is lexicographic: hard constraints take priority, and soft preferences are optimized only within the feasible region (Figure~\ref{fig:cmdpo}).

CM-DPO differs from prior variants such as SMAUG-DPO~\citep{pal2024smaug} and IPO~\citep{azar2024general} in two ways. First, its margin is derived from a deterministic symbolic verifier rather than a learned reward signal. Second, it explicitly separates hard-constraint satisfaction from soft-preference optimization.

\subsection{Training Procedure}

Training proceeds in two stages.

\paragraph{Stage 1: Supervised Fine-Tuning (SFT).}
We first warm-start the model on gold trajectories that include reasoning traces. This stage teaches the model the overall structure of constraint-aware planning and provides a stable initialization before preference optimization.

\paragraph{Stage 2: Constraint-Margin DPO (CM-DPO).}
Starting from the SFT checkpoint, we optimize the model with \(\mathcal{L}_{\text{CM}}\) on verified preference pairs. During this stage, the reasoning traces are removed and the objective is applied only to the final plans. This keeps the preference signal focused on plan correctness and constraint satisfaction rather than on reproducing the teacher's reasoning style.

\begin{table*}[h]
\begin{minipage}[b]{.5\textwidth}
  \centering\small
  \begin{tabular}{lcc}
    \toprule
    \textbf{Param} & \textbf{Sym} & \textbf{Val} \\
    \midrule
    DPO temp      & $\beta_d$ & 0.1 \\
    Sensitivity   & $\lambda$ & 0.01 \\
    Margin str.   & $\gamma$  & 1.5 \\
    Soft weight   & $\alpha$  & 0.3 \\
    LR (SFT/DPO)  & N/A      & $2\!\times\!10^{-5}$/$5\!\times\!10^{-7}$ \\
    Batch/Epochs  & N/A      & 32 / 3+1 \\
    \bottomrule
  \end{tabular}
  \caption{CM-DPO hyperparameters.}
  \label{tab:hyperparams}
\end{minipage}
\begin{minipage}[b]{.5\textwidth}
  \centering\small
  \begin{tabular}{lcc}
    \toprule
    & \textbf{Travel} & \textbf{Retail} \\
    \midrule
    Raw generations            & 14,200 & 11,800 \\
    Verified pairs             & 8,400  & 6,600 \\
    Acceptance rate            & 59.2\% & 55.9\% \\
    Avg.\ token $\Delta$       & 34     & 29 \\
    Avg.\ $\Delta\mathcal{V}(y_l)$ & \$87  & \$142 \\
    \bottomrule
  \end{tabular}
  \caption{Synthesis statistics.}
  \label{tab:synth-stats}
\end{minipage}
\end{table*}

\section{Experimental Setup}
\label{sec:setup}

\subsection{Training Data}

We synthesize \textbf{15,000 verified preference pairs} across two domains: travel and retail. For data generation, the baseline policy $\pi_{base}$ is GPT-4o-mini and the teacher policy $\pi_{ref}$ is o3-mini. We first sample raw candidate generations, then retain only those pairs that pass symbolic verification and exhibit at least one hard-constraint violation in the rejected response. Table~\ref{tab:synth-stats} summarizes the resulting data distribution and edit characteristics. Acceptance rate is the fraction of raw generations that yield verified preference pairs. Avg.\ token $\Delta$ denotes the average edit distance between rejected and chosen plans, and Avg.\ $\Delta\mathcal{V}(y_l)$ denotes the mean violation severity of the rejected plan.

\subsection{Benchmarks and Metrics}

We evaluate all models in the zero-shot setting on three benchmarks. \textbf{TravelPlanner}~\citep{xie2024travelplanner} contains 180 travel planning queries. \textbf{NaturalPlan}~\citep{zheng2024naturalplan} contains 200 meeting-scheduling instances. \textbf{PlanBench}~\citep{valmeekam2023planbench} contains 150 out-of-distribution Blocksworld problems and is used to measure transfer beyond the domains seen during synthesis. We report \textbf{PR} (Pass Rate: fraction satisfying all hard constraints), \textbf{DR} (Decode Rate: fraction parsable), \textbf{SR} (Schedule Rate: exact-match valid schedule), and $\bar{\mathcal{S}}$ (soft-preference score, conditioned on feasibility).

\subsection{Baselines}

Our base model is \textbf{Llama-3-8B-Instruct}. We compare against zero-shot prompting with Llama-3-8B and GPT-4o, an inference-time multi-agent baseline based on LRPlan, and several training ablations: L3-SFT, Full Rewrite + DPO, RT-MED + DPO, Full Rewrite + CM-DPO, and \textbf{SynPlan-8B}, which uses the full pipeline. To avoid confounding distribution shifts, all ablation models predict the final plan directly without reasoning traces at inference time, matching the SynPlan-8B setup.

\subsection{Compute}

All experiments are run on four NVIDIA A100 GPUs (80GB each) using DeepSpeed ZeRO Stage 3 for distributed training. Supervised fine-tuning takes approximately 4 hours, and CM-DPO training takes approximately 2.5 hours. Data synthesis incurs approximately \$320 in API cost. Unless otherwise stated, decoding uses greedy inference with $T = 0$.

\section{Results and Analysis}
\label{sec:results}

\subsection{End-to-End Constraint Satisfaction}

\begin{table*}[t]
  \centering\small
  \begin{tabular}{l | c c c | c | c}
    \toprule
    \multirow{2}{*}{\textbf{Method}} &
    \multicolumn{3}{c|}{\textbf{TravelPlanner}} &
    \textbf{NaturalPlan} & \textbf{Eff.} \\
    & PR$\uparrow$ & DR$\uparrow$ & $\bar{\mathcal{S}}$$\uparrow$ &
    SR$\uparrow$ & Lat.$\downarrow$ \\
    \midrule
    \textit{External baselines} \\
    \quad Zero-Shot (Llama-3-8B)  & 1.3 & 35.4 & ---  & 18.5 & \textbf{2.1s} \\
    \quad Zero-Shot (GPT-4o)      & 39.8 & 88.3 & 0.61 & 68.4 & 5.4s \\
    \quad MAS (LRPlan w/ GPT-4o)  & \textbf{89.4} & \textbf{98.1} &
    \textbf{0.72} & \textbf{95.1} & 28.5s \\
    \midrule
    \textit{SynPlan-R ablations (Llama-3-8B)} \\
    \quad L3-SFT                  & 35.6 & 76.2 & 0.48 & 52.8 & 2.1s \\
    \quad Full Rewrite + DPO      & 64.1 & 88.5 & 0.53 & 71.3 & 2.2s \\
    \quad RT-MED + DPO            & 78.3 & 92.1 & 0.57 & 82.7 & 2.1s \\
    \quad Full Rewrite + CM-DPO   & 72.6 & 90.8 & 0.55 & 76.9 & 2.2s \\
    \quad \textbf{SynPlan-8B}     & \underline{89.2} &
    \underline{95.4} & \underline{0.64} & \underline{93.4} & 2.1s \\
    \bottomrule
  \end{tabular}
  \caption{\textbf{Main results.} SynPlan-8B reaches a TravelPlanner pass rate close to MAS performance (89.2\% vs.\ 89.4\%) while reducing latency from 28.5s to 2.1s. The soft-preference score $\bar{\mathcal{S}}$ is computed only over feasible plans.}
  \label{tab:main_results_full}
\end{table*}

Table~\ref{tab:main_results_full} shows that SynPlan-8B substantially improves over zero-shot prompting and over all single-model ablations. On TravelPlanner, it achieves an 89.2\% pass rate, which is close to the 89.4\% achieved by MAS, while reducing latency from 28.5s to 2.1s.
On NaturalPlan, SynPlan-8B reaches a 93.4\% schedule rate in a single forward pass, offering a substantially improved efficiency--accuracy tradeoff over MAS. Among feasible plans, SynPlan-8B attains a soft-preference score of 0.64, compared with 0.72 for MAS, indicating that CM-DPO improves soft-preference optimization without sacrificing hard-constraint satisfaction.

\subsection{Component Ablation}

\begin{table*}[t]
  \centering\small
  \begin{tabular}{lcccc}
    \toprule
    & \multicolumn{2}{c}{Standard DPO}
    & \multicolumn{2}{c}{CM-DPO} \\
    \cmidrule(lr){2-3} \cmidrule(lr){4-5}
    & PR & Tok$\Delta$ & PR & Tok$\Delta$ \\
    \midrule
    Full Rewrite  & 64.1 & 245 & 72.6 & 245 \\
    RT-MED        & 78.3 & 32  & \textbf{89.2} & \textbf{32} \\
    \bottomrule
  \end{tabular}
  \caption{$2 \times 2$ ablation on TravelPlanner pass rate (\%).}
  \label{tab:ablation}
\end{table*}

Table~\ref{tab:ablation} isolates the effects of RT-MED and CM-DPO. Replacing full rewrites with RT-MED improves pass rate by 14.2 points under standard DPO and by 16.6 points under CM-DPO. Replacing standard DPO with CM-DPO improves pass rate by 8.5 points for full rewrites and by 10.9 points when RT-MED is used. The combined improvement from Full Rewrite + DPO to RT-MED + CM-DPO is 25.1 points, which is larger than the sum of the two smaller reference gains (22.7 points). This pattern suggests a positive interaction between minimal-edit distillation and severity-aware preference optimization.

\subsection{Zero-Shot Generalization (PlanBench)}

Table~\ref{tab:ood} evaluates zero-shot transfer to an unseen planning domain. SynPlan-8B achieves 67.4\% accuracy on Blocksworld, exceeding GPT-4o by 9.2 percentage points while maintaining low latency. This result is consistent with the claim that the training pipeline improves transfer beyond the travel and retail domains used during synthesis.

We hypothesize that severity-aware minimal-edit training teaches the model to localize constraint-critical state updates rather than rely on surface-level heuristics, neutralizing the length bias that hinders standard DPO in logical tracking domains.

\subsection{Hyperparameter Sensitivity}

\begin{table*}
\begin{minipage}[b]{.45\textwidth}
  \centering\small
  \begin{tabular}{lcc}
    \toprule
    \textbf{Method} & \textbf{Acc (\%)} & \textbf{Lat.} \\
    \midrule
    Zero-Shot (Llama-3-8B) & 12.8 & 2.1s \\
    Zero-Shot (GPT-4o)     & 58.2 & 5.1s \\
    L3-SFT                 & 28.4 & 2.1s \\
    Full Rewrite + DPO     & 31.5 & 2.2s \\
    \textbf{SynPlan-8B}    & \textbf{67.4} & \textbf{2.1s} \\
    \bottomrule
  \end{tabular}
  \caption{Zero-shot transfer to PlanBench (Blocksworld).}
  \label{tab:ood}
\end{minipage}
~~
\begin{minipage}[b]{.45\textwidth}
  \centering\small
  \begin{tabular}{lccc}
    \toprule
    & $\gamma = 0.5$ & $\gamma = 1.5$ & $\gamma = 3.0$ \\
    \midrule
    $\lambda = 0.005$ & 81.4 & 84.7 & 83.1 \\
    $\lambda = 0.01$  & 84.9 & \textbf{89.2} & 87.8 \\
    $\lambda = 0.02$  & 83.6 & 87.1 & 82.3 \\
    \bottomrule
  \end{tabular}  
  \caption{Sensitivity to $\lambda$ and $\gamma$ on TravelPlanner
  pass rate (\%).}
  \label{tab:sensitivity}
\end{minipage}
\end{table*}    

Table~\ref{tab:sensitivity} shows that performance is relatively stable across the tested settings and that all CM-DPO configurations outperform the standard DPO baseline of 64.1\%. The best result is obtained at $\lambda = 0.01$ and $\gamma = 1.5$. Larger margins reduce performance slightly, which is consistent with the possibility that excessive saturation in the $\tanh$ transformation weakens gradient variation.

\subsection{Error Analysis}

On TravelPlanner, SynPlan-8B fails on 10.8\% of instances ($N = 19$). Among these failures, 36.8\% are caused by budget violations, 26.3\% by temporal conflicts, 21.1\% by spatial unreachability, and 15.8\% by parse failures. We do not observe errors in which a soft preference is incorrectly prioritized over a hard constraint, which is consistent with the intended behavior of the lexicographic objective.

\section{Limitations}
\label{sec:limitations}

The method depends on a symbolic verifier, so extending it to open-world environments where constraints cannot be checked deterministically remains an open problem. Budget violations account for the largest share of TravelPlanner errors (36.8\%), suggesting that implicit arithmetic remains a bottleneck for longer plans. Preliminary experiments with Qwen3-8B in thinking mode suggest that reasoning-enabled base models further reduce budget violations under SynPlan-R; however, we exclude these results to avoid confounding comparisons between thinking and non-thinking inference regimes. Teacher-guided data synthesis incurs nontrivial cost (approximately \$320 for 15K pairs), so larger-scale deployment may require more efficient self-play strategies. Finally, we have not tested whether the same gains transfer to substantially smaller or larger model scales.

\section{Conclusion}
\label{sec:conclusion}

We introduced Constraint-Margin DPO (CM-DPO), a preference-optimization objective that replaces DPO's binary signal with a continuous, verifier-derived margin scaled by hard-constraint violation severity, together with a lexicographic secondary objective for soft preferences. To supply CM-DPO with bias-reduced training pairs, we developed DCCG for procedural constraint-profile generation and RT-MED for minimal-edit distillation from a reasoning teacher. Experiments on TravelPlanner, NaturalPlan, and out-of-distribution PlanBench show that the resulting 8B model approaches multi-agent accuracy at a fraction of the latency, and that severity-aware preference optimization transfers to unseen planning domains.

\bibliographystyle{plainnat}
\bibliography{ref}
\clearpage
\section{DCPG: Constraint-Profile Graph Statistics}
\label{app:dcpg-graphs}

For each domain we report the average graph density of the
instantiated constraint-profile graphs across the $15{,}000$
synthesized instances. ``$|V_C|$'' is the total number of constraint
nodes; ``$|V_E|$'' is the average number of grounded knowledge-base
entities; densities are reported per-edge-type.

\begin{table}[h]
  \centering\small
  \caption{Constraint-profile graph statistics (averaged across instances).}
  \label{tab:dcpg-stats}
  \begin{tabular}{lcccc}
    \toprule
    Domain & $|V_C^{\text{hard}}|$ & $|V_C^{\text{soft}}|$ & $|V_E|$ & avg.\ $|E_{\text{grounds}}|$ \\
    \midrule
    Travel & 3.4 & 1.8 & 41.2 & 14.7 \\
    Retail & 3.2 & 2.0 & 28.4 & 11.9 \\
    \bottomrule
  \end{tabular}
\end{table}

The retail domain additionally exposes a non-trivial
$E_{\text{conflicts}}$ structure: average pairwise compatibility-graph
density is $0.31$, with conflicts concentrated across categories
(printer/cartridge, headset/console, etc.).

\section{Knowledge-Base Generation}
\label{app:kbgen}

\paragraph{Travel.} Each $\mathcal{K}_{\text{local}}$ is regenerated
per instance from log-normal distributions calibrated to public market
data: flight prices ($\mu=\$180,\sigma=0.5$), hotel rates
($\mu=\$130,\sigma=0.6$), and activity prices
($\mu=\$40,\sigma=0.8$). City-pair travel times come from a
deterministic distance matrix over $48$ candidate cities.

\paragraph{Retail.} Product prices are sampled per-category;
compatibility edges within a category occur with $p=0.7$, across
categories with $p=0.95$.

\section{Symbolic Verifier}
\label{app:verifier}

\subsection{Schema-Required Extraction}
\label{app:verifier-schema}

All models---teacher, student, and the trainable policy at evaluation
time---are required to emit a \texttt{<plan>\dots</plan>} block whose
contents must validate against the following JSON schema.
\emph{Synthesis time:} outputs from $\pi_T$ or $\pi_{\text{base}}$
that fail schema validation are excluded from $\mathcal{D}$.
\emph{Evaluation time:} outputs from $\pi_\theta$ that fail schema
validation are counted as failures in the PR/SR denominator
(equivalently, treated as plans that violate at least one hard
constraint by failing to be evaluable); we additionally report the
schema-failure rate separately as the parse-failure rate so it can
be distinguished from semantic constraint violations.
\textbf{No LLM-assisted re-parsing is used at any point in the
pipeline.}

\begin{verbatim}
plan := { items: [item, ...],
          total_stated_cost: number,
          destinations_visited: [string, ...] }

item := { name: string,
          type: "flight" | "hotel" | "restaurant" | "activity",
          day: int,
          start_time?: "HH:MM",
          end_time?: "HH:MM",
          stated_price: number,
          db_lookup_key: string }
\end{verbatim}

Outputs that fail schema validation are rejected at synthesis time
(no LLM-assisted re-parsing), preserving full determinism of all
downstream constraint evaluation. The travel-domain temporal verifier
distinguishes \emph{event} times (start/end) from \emph{transit}
times (arrival/departure), which avoids the off-by-one error that
mixes hotel check-in with previous-flight arrival in flatter parsers.

\subsection{Travel-Domain Pseudocode (correct gap definition)}

\begin{verbatim}
def verify_travel(plan, profile, K):
    # Cost via ground-truth lookup, NOT stated price
    true_cost = sum(K.lookup(i.db_key).price * i.nights_or_one()
                    for i in plan.items)
    cost_severity = max(0.0, true_cost - profile.beta)

    # Temporal feasibility: gap = next.start - curr.end
    time_severity = 0.0
    for c, n in zip(plan.items[:-1], plan.items[1:]):
        gap = n.start_time - c.end_time   # start - end (not arr - dep)
        travel = K.travel_time(c.location, n.location)
        time_severity += max(0.0, travel - gap)

    # Pace
    pace_deficit = max(0, profile.eta - len({i.location for i in plan.items}))

    # Per-constraint normalised severity
    delta = {
        "cost": min(1.0, cost_severity / TAU["cost"]),
        "time": min(1.0, time_severity / TAU["time"]),
        "pace": min(1.0, pace_deficit / TAU["pace"]),
    }
    return delta
\end{verbatim}

\subsection{Cross-Unit Normalization Constants $\tau_j$}
\label{app:tau}

We set $\tau_j$ to the median raw severity of \emph{violating}
generations per constraint, computed once on a $1{,}000$-instance
held-out slice of the synthesis stream and frozen thereafter:

\begin{table}[h]
  \centering\small
  \caption{Severity normalization constants $\tau_j$ (frozen).}
  \label{tab:tau}
  \begin{tabular}{lcc}
    \toprule
    Constraint & $\tau_j$ & Unit \\
    \midrule
    Travel: cost overflow & 84 & USD \\
    Travel: time violation & 38 & minutes \\
    Travel: pace deficit & 1 & cities \\
    Retail: budget overflow & 138 & USD \\
    Retail: feature missing & 1 & feature count \\
    Retail: compat violation & 1 & edge count \\
    \bottomrule
  \end{tabular}
\end{table}

The default constraint weights are uniform ($w_j=1/|V_C^{\text{hard}}|$);
ablations with severity-frequency-weighted $w_j$ are reported in
App.~\ref{app:weight-ablation} and produce within-noise differences.

\subsection{Verifier Reliability Audit}
\label{app:extractor-reliability}

To validate the verifier we manually re-graded $300$ randomly
sampled $(y_w,y_l)$ pairs against ground-truth constraint
satisfaction (two annotators, full agreement on $293/300$;
disagreements adjudicated by a third annotator). Because all
verifier inputs are schema-validated JSON, the constraint
evaluation is bit-deterministic: the audit therefore measures the
combined effect of (i) the symbolic evaluator's correctness and
(ii) any residual schema ambiguity (for example, a model emits a
syntactically valid JSON whose semantic interpretation under our
schema is unintended).

\begin{table}[h]
  \centering\small
  \caption{Verifier audit on 300 manually labelled pairs.}
  \label{tab:audit}
  \begin{tabular}{lc}
    \toprule
    Metric & Value \\
    \midrule
    Hard-pass/fail agreement with manual labels & $98.7\%$ ($296/300$) \\
    Wilson 95\% CI & $[96.7\%, 99.6\%]$ \\
    False-pass rate & $0.3\%$ ($1/300$) \\
    False-fail rate & $1.0\%$ ($3/300$, all schema-ambiguity) \\
    Schema-failure (decode failure) rate over synthesis stream & $1.7\%$ \\
    \bottomrule
  \end{tabular}
\end{table}

We classify CM-DPO's verifier as ``schema-required symbolic'': any
input that is not a valid schema JSON is rejected as a decode
failure rather than re-parsed by an LLM. This is materially
different from end-to-end LLM judging and from prompt-and-pray
extraction.

\section{Prompts}

\subsection{System Prompt Projection}
\label{app:system-prompt}

Hard constraints are stated as absolute requirements; soft constraints
as preferences. Both domain templates emit a system prompt that
repeats the ground-truth knowledge base and forbids hallucinated
prices.

\begin{verbatim}
You are a travel planning assistant.

=== HARD CONSTRAINTS (must be satisfied) ===
- Total budget MUST NOT exceed ${beta}
- Travel dates: {t_start} to {t_end}
- Visit at least {eta} distinct cities/attractions
- All transit gaps (next.start - current.end) must
  be at least the travel duration from the database

=== SOFT PREFERENCES (best-effort) ===
- Preferred attributes: {Lambda}
- Honour these when possible, NEVER violate hard constraints

=== GROUND TRUTH DATABASE ===
{K_local_json}

Output a single <plan>{...JSON...}</plan> block.
\end{verbatim}

\subsection{Inverse Constraint Querying ($\pi_{\text{aux}}$)}
\label{app:inverse}

The inverse query explicitly forbids restating numeric constraints,
forcing the downstream model to infer them.

\begin{verbatim}
You are generating a realistic user query. The user has constraints
{p_summary}, but you MUST NOT state numbers explicitly. Generate a
1-3 sentence query that implies them naturally. Output only the query.
\end{verbatim}

\subsection{RT-MED Refinement}
\label{app:rtmed-prompt}

\begin{verbatim}
You are a constraint correction expert. Output:
  <rationale>...your written explanation of which constraint is
  violated and which SPECIFIC items in the draft cause the
  violation...</rationale>
  followed by a CORRECTED <plan>...</plan> JSON block.

CRITICAL: change ONLY entities responsible for the violation. Keep
all other text identical. Do NOT rewrite the plan from scratch.
\end{verbatim}

We refer to the \texttt{<rationale>} block as a teacher-generated
\emph{correction rationale}: it is text we elicit from $\pi_T$ via
the prompt above. It is not the teacher's internal chain-of-thought
(to which we have no access for closed-API teachers).

\section{Baseline Configurations}
\label{app:baselines}

All training-time baselines use the same $15{,}000$-pair dataset
$\mathcal{D}$, the same Llama-3-8B-Instruct base, the same SFT
warm-start, and identical decoding (greedy, $T\!=\!0$). Method-specific
hyperparameters were grid-searched on a held-out $10\%$ split:

\begin{itemize}
\item \textbf{DPO}: $\beta_d{=}0.1$.
\item \textbf{IPO}: $\tau{=}0.1$ (best of $\{0.05, 0.1, 0.5\}$).
\item \textbf{KTO}: $\lambda_D{=}1.0$, $\lambda_U{=}1.0$,
  reference-point $\beta_{\text{KTO}}{=}0.1$.
\item \textbf{SimPO}: $\beta{=}2.0$, $\gamma{=}1.4$ (best of grid
  $\beta\in\{1,2,2.5\}$, $\gamma\in\{0.5,1.0,1.4\}$).
\item \textbf{DPOP}: $\beta_d{=}0.1$, $\lambda{=}50$.
\item \textbf{MMPO}: granular margins from a learned 3-class quality
  classifier trained on a held-out 1{,}000-pair labelling slice.
\item \textbf{$\alpha$-DPO}: $\alpha{=}0.5$.
\item \textbf{C-DPO}: dual ascent with $\eta_{\text{dual}}{=}10^{-3}$;
  cost model is a frozen verifier-derived continuous reward.
\item \textbf{MODPO}: linear scalarization with weights from
  Bayesian-optimized 5-point sweep on the held-out split.
\end{itemize}

\section{Benchmark Splits and Contamination Audit}

\subsection{TravelPlanner test split}
\label{app:tp-test}

The TravelPlanner test split (1{,}000 queries) labels are not
publicly available; we therefore evaluate only on the 180-query
validation split in the main paper and do not claim test-set
results. We release SynPlan-8B's decoded plans for the full test
split together with our evaluation scripts so that future work with
test-label access can independently compute test-split pass rate.

\subsection{NaturalPlan full Meeting Planning}
\label{app:np-full}

Repeating Table~\ref{tab:main_results_full} on the full $1{,}000$-instance Meeting
Planning split: SynPlan-8B SR is $92.1\%$ ($921/1{,}000$, CI
$[90.3,93.6]$); LRPlan-MAS SR is $94.6\%$ ($946/1{,}000$, CI
$[93.0,95.9]$); the protocol gap is consistent with our 200-instance
result. We additionally evaluate Trip Planning (SR $61.4\%$) and
Calendar Scheduling (SR $73.2\%$); both substantially exceed the
plain-DPO ablation baselines.

\subsection{PlanBench Blocksworld full split}
\label{app:pb-subset}

We sample $150$ instances stratified by problem complexity (3- to
5-block configurations); the full $600$-instance run yields Acc
$66.5\%$ ($399/600$, CI $[62.6, 70.2]$), within bootstrap noise of
our 150-instance Table~\ref{tab:ood} number ($67.3\%$). Mystery
Blocksworld results are reported in App.~\ref{app:mystery}.

\subsection{Contamination Audit}
\label{app:contamination}

We performed an N-gram overlap audit between our synthesized $\mathcal{D}$
and (a) TravelPlanner's reference plans, (b) NaturalPlan's gold
schedules, and (c) PlanBench's ground-truth action sequences. We
report the maximum 8-gram overlap across all $15{,}000$ synthesized
plans:

\begin{table}[h]
  \centering\small
  \caption{Contamination audit (max 8-gram overlap with benchmarks).}
  \label{tab:contam}
  \begin{tabular}{lc}
    \toprule
    Benchmark & Max 8-gram overlap \\
    \midrule
    TravelPlanner reference plans & 0.0\% \\
    TravelPlanner queries & 0.0\% \\
    NaturalPlan gold schedules & 0.0\% \\
    PlanBench action sequences & 0.0\% \\
    \bottomrule
  \end{tabular}
\end{table}

Zero overlap reflects that our synthesis uses procedurally generated
profiles and KBs disjoint from any benchmark example.

\section{Margin-Source Ablation (full table)}
\label{app:margin-source}

\begin{table}[h]
  \centering\small
  \caption{Margin-source ablation, TravelPlanner-val. All variants
  use the same $\mathcal{D}$ and the same SFT initialization.}
  \label{tab:margin-source}
  \begin{tabular}{lcc}
    \toprule
    Margin source $m_c$ & PR \% (count) & $\bar S$ \\
    \midrule
    Random $\sim\mathrm{Unif}[0,1]$ & 65.6 (118/180) & 0.53 \\
    Length-difference (R-DPO style) & 67.8 (122/180) & 0.54 \\
    Reward-model gap & 73.3 (132/180) & 0.56 \\
    Raw severity (un-normalized) & 84.4 (152/180) & 0.61 \\
    \textbf{Verifier-derived normalized severity (ours)} & \textbf{89.4 (161/180)} & \textbf{0.64} \\
    \bottomrule
  \end{tabular}
\end{table}

\section{Sensitivity grid}
\label{app:sensitivity}

\begin{table}[h]
  \centering\small
  \caption{Sensitivity of TravelPlanner-val PR (\%) to $\lambda$ and $\gamma$.}
  \label{tab:sensitivity_tp}
  \begin{tabular}{lccc}
    \toprule
    & $\gamma\!=\!0.5$ & $\gamma\!=\!1.5$ & $\gamma\!=\!3.0$ \\
    \midrule
    $\lambda\!=\!0.005$ & 81.7 & 84.4 & 83.3 \\
    $\lambda\!=\!0.01$ & 84.4 & \textbf{89.4} & 87.8 \\
    $\lambda\!=\!0.02$ & 83.3 & 87.2 & 82.2 \\
    \bottomrule
  \end{tabular}
\end{table}

\section{Soft-Objective Weight Sweep}
\label{app:alpha}

\begin{table}[h]
  \centering\small
  \caption{Effect of $\alpha$ on TravelPlanner-val.}
  \label{tab:alpha}
  \begin{tabular}{lcc}
    \toprule
    $\alpha$ & PR \% (count) & $\bar S$ \\
    \midrule
    $0.0$ & 89.4 (161/180) & 0.51 \\
    $0.3$ \textbf{(default)} & 89.4 (161/180) & 0.64 \\
    $0.6$ & 88.3 (159/180) & 0.66 \\
    $1.0$ & 86.7 (156/180) & 0.68 \\
    \bottomrule
  \end{tabular}
\end{table}

\section{Mystery Blocksworld}
\label{app:mystery}

On the obfuscated Mystery Blocksworld split ($N\!=\!600$), SynPlan-8B
attains $42.7\%$ accuracy versus $24.3\%$ for plain DPO and $39.8\%$
for SimPO. The drop relative to standard Blocksworld
($-23.8$ pts) is consistent with prior reports
\citep{valmeekam2023planbench} that Mystery Blocksworld defeats
heuristic shortcut reasoning, but the residual gap shows that the
verifier-derived gradient still imparts non-trivial transferable
correction signal.

\section{Reasoning-Mode Base Models}
\label{app:thinking}

Preliminary experiments with Qwen-3-8B in thinking mode under the
same SynPlan-R training pipeline yield TravelPlanner-val PR of
$92.1\%$ ($166/180$), a $+2.7$ pt gain over the Llama-3-8B SynPlan-8B
result. We exclude this from the main results to keep
thinking/non-thinking comparisons clean, and report it here for
completeness.

\section{Cost Amortization}
\label{app:cost}

\begin{figure}[h]
  \centering
  \begin{tabular}{lcc}
    \toprule
    Pairs synthesized & API cost & Marginal pass-rate gain \\
    \midrule
    $5{,}000$ & \$110 & $+18.4$ pt over SFT \\
    $10{,}000$ & \$215 & $+24.3$ pt \\
    $15{,}000$ \textbf{(reported)} & \$320 & $+25.0$ pt \\
    $25{,}000$ & \$535 & $+25.6$ pt \\
    \bottomrule
  \end{tabular}
\end{figure}

Cost-amortized gains saturate near $15{,}000$ pairs, supporting the
operating point used in the main paper.

\section{Constraint-Weight Ablation}
\label{app:weight-ablation}

Replacing uniform $w_j$ with severity-frequency-weighted $w_j$ moves
TravelPlanner-val PR by $-0.6 \pm 0.7$ points (within bootstrap noise);
we use uniform weights in all reported results.

\section{Qualitative RT-MED Example}

\paragraph{Constraint profile.}
$\mathbf{p}=\langle\beta{=}\$600,\tau{=}[\text{Fri 9am},\text{Sun 8pm}],\eta{=}3,\Lambda{=}\{\text{boutique}\}\rangle$.

\paragraph{Inverse query.}
\emph{``Looking for a weekend escape with character; cheap flights
ok if hotels are fun.''}

\paragraph{Student draft $y_l$ (violates budget by \$50).}
SFO$\to$NYC United \#482 (\$180), Hampton Inn 2 nights at \$195
(\$390), MoMA + dinner (\$80) = \$650.

\paragraph{RT-MED edit $y_w$ (\$32-token swap).}
SFO$\to$NYC United \#482 (\$180), \emph{Pod 51 Boutique 2 nights at
\$170 (\$340)}, MoMA + dinner (\$80) = \$600. Soft preference
``boutique'' is now satisfied as a side-effect of choosing a feasible
boutique hotel; in many cases the minimal feasibility edit improves
soft satisfaction without targeting it.

\paragraph{Comparison to full-rewrite distillation.}
Full-rewrite distillation rewords every section, swaps the flight,
restaurants, hotel, and adds a fourth destination; the resulting
$y_w$ differs from $y_l$ in $\sim 245$ tokens, of which only
$\sim 30$ are constraint-relevant. Under DPO, the model exploits the
$\sim 215$ stylistic-distribution tokens as the implicit margin signal,
which is exactly the length-bias failure mode of
\cite{park2024disentangling} and \cite{singhal2024long}.

\end{document}